\documentclass{article}

\usepackage{microtype}
\usepackage{graphicx}
\usepackage{subcaption}
\usepackage{booktabs} 

\usepackage{hyperref}

\usepackage[preprint]{icml2026}

\usepackage{amsmath}
\usepackage{amssymb}
\usepackage{mathtools}
\usepackage{amsthm}

\RequirePackage{xspace}
\makeatletter
\DeclareRobustCommand\onedot{\futurelet\@let@token\@onedot}
\def\@onedot{\ifx\@let@token.\else.\null\fi\xspace}

\def\etal{\emph{et al}\onedot}
\usepackage{multirow}
\usepackage{pifont}
\usepackage{csquotes}

\usepackage[capitalize,noabbrev]{cleveref}

\theoremstyle{plain}

\theoremstyle{definition}

\theoremstyle{remark}

\usepackage[textsize=tiny]{todonotes}

\icmltitlerunning{SSR: A Training-Free Approach for Streaming 3D Reconstruction}

\begin{document}

\twocolumn[
  \icmltitle{Self-Expressive Sequence Regularization: \\A Training-Free Approach for Streaming 3D Reconstruction}



  \icmlsetsymbol{equal}{*}

  \begin{icmlauthorlist}
    \icmlauthor{Hui Deng}{NWPU}
    \icmlauthor{Yuxin Mao}{NWPU}
    \icmlauthor{Yuxin He}{NWPU}
    \icmlauthor{Yuchao Dai}{NWPU}
  \end{icmlauthorlist}

  \icmlaffiliation{NWPU}{Shool of Electronics and Information, Northwestern Polytechnical University and Shaanxi Key Laboratory of Information Acquisition and Processing, Xi’an, China.}

  \icmlcorrespondingauthor{Hui Deng}{denghui986@mail.nwpu.edu.cn}
  \icmlcorrespondingauthor{Yuchao Dai}{daiyuchao@nwpu.edu.cn}

  \icmlkeywords{Self-expression, 3D vision}

  \vskip 0.3in
]



\printAffiliationsAndNotice{}  
\begin{abstract}

    Streaming 3D reconstruction demands long-horizon state updates under strict latency constraints, yet stateful recurrent models often suffer from geometric drift as errors accumulate over time.
    We revisit this problem from a Grassmannian manifold perspective: the latent persistent state can be viewed as a subspace representation, i.e., a point evolving on a Grassmannian manifold, where temporal coherence implies the state trajectory should remain on (or near) this manifold.
    Based on this view, we propose \textbf{Self-expressive Sequence Regularization (SSR)}, a plug-and-play, \textbf{training-free} operator that enforces Grassmannian sequence regularity during inference.
    Given a window of historical states, SSR computes an analytical affinity matrix via the self-expressive property and uses it to regularize the current update, effectively pulling noisy predictions back toward the manifold-consistent trajectory with minimal overhead.
    Experiments on long-sequence benchmarks demonstrate that SSR consistently reduces drift and improves reconstruction quality across multiple streaming 3D reconstruction tasks.
\end{abstract}
\section{Introduction}

Humans possess the remarkable ability to construct and refine a coherent 3D mental model of their environment from a continuous stream of visual observations. In computer vision, this capability is pursued through streaming 3D reconstruction~\cite{wu2025point3r, wang2025continuous}, which aims to estimate camera poses and dense geometry in real-time. While offline methods~\cite{wang2023dust3r, wang2025vggt} leveraging global alignment achieve high accuracy, they suffer from quadratic computational growth and cannot process sequences of arbitrary length. In contrast, state recurrent models, such as CUT3R~\cite{wang2025continuous}, maintain a constant-sized persistent state to encode scene context, offering low complexity and high-speed inference.

Despite their efficiency, state recurrent models are plagued by geometric drift and catastrophic forgetting. As the sequence length exceeds the training context, the recurrent update mechanism often fails to reconcile new observations with historical context, leading to broken trajectories and distorted pointmaps. Recent work attempts to mitigate this, TTT3R~\cite{chen2025ttt3r} deems frame reconstruction as an online learning problem, Long3r~\cite{chen2025long3r} designed a spatio-temporal unified feature to preserve long-range contextual information. Although these approaches are highly innovative, we believe they have yet to fully leverage the inherent properties of sequence structure. Such properties enable trained models to maintain information consistency across long-range data, thereby reducing reconstruction drift and enhancing overall performance.



In this work, we cast streaming 3D reconstruction as the problem of maintaining a
\emph{trajectory on a Grassmannian manifold}, rather than simply updating a hidden memory.
Specifically, each latent state summarizes the scene using a low-dimensional subspace representation and thus corresponds to a point on the Grassmannian, i.e., the space of $k$-dimensional subspaces.
Under this formulation, temporal coherence naturally imposes a geometric constraint: the latent states should evolve smoothly over time and remain on, or close to, the underlying manifold.
This perspective is further motivated by classical Non-Rigid Structure-from-Motion (NRSfM), where coherent shape deformations or smoothly varying viewpoints are known to admit low-rank subspace structures that evolve along a Grassmannian manifold~\cite{kumar2018scalable}.



A key consequence of this viewpoint is the \emph{self-expressive} property on the manifold: a state in a coherent sequence can be expressed as a linear combination of other states within the same sequence.
Let $\mathbf{S} = [\mathbf S_1, \dots, \mathbf S_t]^\top \in \mathbb{R}^{t \times d}$ denote the latent state sequence. Self-expressive implies the existence of an affinity matrix $\mathbf{C} \in \mathbb{R}^{t \times t}$ such that $\mathbf{S} = \mathbf{C}\mathbf{S}$.


In a streaming context, the current predicted state $\mathbf S_t$ may \emph{drift off} the Grassmannian due to occlusion, low texture, or poor parallax. By computing an analytical affinity matrix $\mathbf{C}$ from similarities between the current observation and a historical window, we obtain a closed-form, training-free way to pull the state back toward the self-expressive Grassmannian structure, thereby reducing accumulated drift.


We introduce \textbf{Self-expressive Sequence Regularization (SSR)}, a training-free, plug-and-play operator that enforces Grassmannian sequence regularity during the inference time.
SSR uses the \textbf{state} sequence within a historical window to compute $\mathbf{C}$ and regularize the update process so that the latent trajectory better respects the self-expressive constraint.
This operation performs a temporal ``consensus'' on the manifold, using the collective information of nearby states to suppress geometric inconsistencies in the current frame.
By imposing explicit Grassmannian structure on the latent updates of 3D foundation models, we can achieve an efficient and structurally grounded perception system.

Our contributions are summarized as follows:
\begin{itemize}
    \item \textbf{A Grassmannian View of State Persistence:} We connect NRSfM subspace theory to streaming 3D reconstruction and interpret the persistent latent state as a point evolving on a Grassmannian manifold, rather than a generic hidden memory.
    \item \textbf{Self-expressive Sequence Regularization (SSR):} We propose an analytical, training-free correction mechanism derived from Grassmannian self-expressive, mitigating drift and forgetting without backpropagation or additional parameters.
    \item \textbf{Empirical Validation:} We validate SSR across many tasks, including video depth estimation and camera pose estimation, and demonstrate consistent improvements on long-sequence and dynamics scenarios.
\end{itemize}

\section{Related Works}

\noindent\textbf{Conventional Reconstruction. } Reconstructing the corresponding 3D structure from a set of ordered or unordered observation sequences has long been a significant challenge in the field of computational vision. From early self-calibrating reconstruction systems~\citep{mohr1995relative,pollefeys2004visual,beardsley19963d,fitzgibbon1998automatic} to large-scale reconstruction systems, 3D reconstruction algorithms have made significant strides. In the early stages of 3D reconstruction, various strategies were explored, including incremental reconstruction~\cite{agarwal2011building,frahm2010building,wu2013towards} and global reconstruction~\cite{wilson2014robust,sweeney2015optimizing}. 
Nowadays, based on factors such as applicability and practical performance, incremental reconstruction has emerged as the most widely adopted and effective solution. Among these, the introduction of COLMAP~\cite{schonberger2016structure} standardized the computational framework for incremental Structure from motion(SfM), with the PnP-centered image registration loop becoming the primary workflow for traditional SfM.  However, SfM can only compute sparse scene point clouds and camera poses. To obtain a denser and more accurate scene model, Multi-View Stereo~\cite{furukawa2015multi} computation must be performed. Inverse rendering-based reconstruction methods have entered the mainstream spotlight with their remarkable expressive power. 
A representative method in this category is NERF~\cite{mildenhall2021nerf}, and NERF-based techniques have gradually expanded into diverse application scenarios, from initial small-scale implementations~\cite{deng2022depth, ma2022deblur, barron2022mip} to subsequent large-scale deployments~\cite{tancik2022block}. Later reconstruction methods centered on Gaussian Splating~\cite{kerbl20233d} also share similar underlying principles.

\noindent\textbf{Learning-based Reconstruction.} With the advancement of deep learning technology, many studies have begun to explore its application in 3D reconstruction. SfMLeaner~\cite{zhou2017unsupervised} pioneered fully end-to-end learning-based reconstruction, followed by numerous subsequent methods~\cite{zhang2023lite, zhao2022monovit} adhering to this paradigm.
Another approaches~\cite{bloesch2018codeslam, czarnowski2020deepfactors} adopted the strategy of integrating depth estimation networks with optimization structure. With the widespread adoption of Transformer~\cite{vaswani2017attention}, fully data-driven end-to-end models regained prominence. Methods like Dust3R~\cite{wang2023dust3r}, utilizing image pairs, enabled networks to autonomously learn spatial-planar geometric relationships, successfully implementing data-driven 3D reconstruction algorithms~\cite{leroy2024grounding,zhang2024monst3r, chen2025easi3r, zhu2025aether}. Approaches such as VGGT~\cite{wang2025vggt} further demonstrated that well-designed network models, when paired with massive datasets, can directly learn geometric knowledge from sequences. However, due to the limited capability of such methods to handle long sequence inputs, reconstruction~\cite{wang20243d, wu2025point3r, wang2025continuous} approaches employing online updates have begun to emerge. This paper aims to explore how to enhance the reconstruction capabilities of streaming reconstruction methods for long-range sequences.

\noindent\textbf{Non-rigid Reconstruction.} The purpose of non-rigid reconstruction is to view dynamic and static objects through a unified perspective. This implies that geometric or physical constraints applicable to specific scenarios cannot be utilized. Consequently, non-rigid reconstruction methods typically opt for solutions based on numerical structure. Bregler~\etal~\cite{bregler2000recovering} first introduced factorization framework~\cite{tomasi1992shape} for Non-rigid Structure-from-Motion. When confronted with an ambiguity crisis~\cite{xiao2004closed}, Akhter~\etal~\cite{akhter2009defense} demonstrated obtaining approximately unique solutions under orthogonal constraints. Subsequent approaches built upon this foundation, proposing different solutions, including using DCT basis to recover motion trajectory~\cite{akhter2008nonrigid, gotardo2011computing}, using low-rank constraint to regularize 3D structure recovery~\cite{paladini2010sequential, dai2014simple, kumar2020non, kumar2022organic}, Novotny \etal~\cite{novotny2019c3dpo} proposed the first deep learning framework for NRSfM. Subsequently, single-frame reconstruction methods~\cite{kong2020deep,wang2020deep,park2020procrustean} designed from different angles emerged.However, when dealing with sequential input, a better approach for achieving superior results is to seek the ability to extract information from context~\cite{zeng2021pr,wang2021paul,zeng2022mhr,choi2019can,sidhu2020neural,deng2022deep}.  
Self-expressive models, serve as a valuable tool for constraining sequence reconstruction. However, their application in other domains remains limited. The proposed method explores how to leverage this tool within streaming reconstruction frameworks.

\section{Methodology}

In this section, 
we introduce Self-expressive Sequence Regularization (SSR), a training-free, plug-and-play operator designed to enforce contextual consistency during inference. We start from a brief introduction to the Grassmannian Manifold in \S~\ref{sec:sec31}, then discuss self-expressive in \S~\ref{sec:sec31} and \S~\ref{sec:sec32}, and finally outline the proposed method in \S~\ref{sec:sec34}.
\begin{figure*}[htbp]
    \centering
    \includegraphics[width=1.0\textwidth]{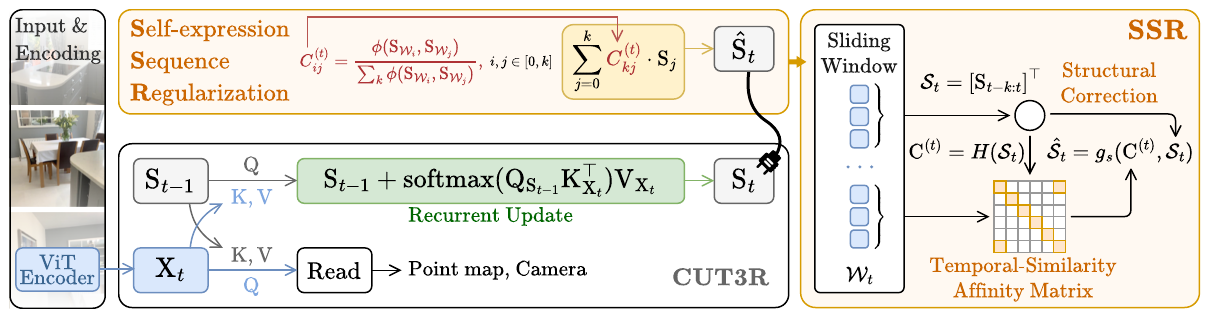} 
    \caption{\textbf{Illustration of the Self-expressive Sequence Regularization (SSR).} We introduce a training-free regularization scheme for CUT3R. Specifically, we refine the frame-wise states calculated by foundation models through a sliding-window reconstruction process. By leveraging the affinity between temporal states, our method achieves sequence regularization without introducing learnable parameters, making it directly applicable to off-the-shelf pre-trained foundation model.}
    \label{fig:pipeline}
\end{figure*}

\subsection{Grassmannian Manifold View}
\label{sec:sec31}
The Grassmann manifold, denoted as $\mathcal{G}(n,r)$ with $n>r$, is the set of all $r$-dimensional linear subspaces of $\mathbb{R}^n$.
A point on $\mathcal{G}(n,r)$ is an equivalence class $[\mathbf U]$ represented by any $n\times r$ matrix $\mathbf U$ whose columns form an orthonormal basis spanning that subspace, i.e., $[\mathbf U] \triangleq \mathrm{span}(\mathbf U)$.

To compare two subspaces $[\mathbf U_1]$ and $[\mathbf U_2]$, we use the \emph{projection metric}:
\begin{equation}
    d_{\mathrm{proj}}\big([\mathbf U_1],[\mathbf U_2]\big) = \frac{1}{\sqrt{2}}\left\| \mathbf U_1\mathbf U_1^{\top} - \mathbf U_2\mathbf U_2^{\top} \right\|_F.
\end{equation}
Which is convenient since it embeds each Grassmann point into a Euclidean space through the mapping $\mathbf U\mapsto \mathbf U\mathbf U^{\top}$. For more information, please refer to \cite{kumar2018scalable}.

With this definition in hand, our key viewpoint is to interpret the persistent latent state in streaming reconstruction as a compact \emph{subspace descriptor} of the scene.
That is, each state corresponds to a point on $\mathcal{G}(n,r)$, and the online recurrent update induces a temporal \emph{trajectory} on the manifold.
This provides a structural prior: for coherent scene evolution, successive states should remain close under $d_{\mathrm{proj}}$ (equivalently, have similar projection matrices), whereas long-horizon drift can be viewed as the state trajectory deviating from this manifold-consistent evolution.

\subsection{Recurrent Streaming Reconstruction}
\label{sec:sec32}

We consider RNN-style streaming reconstruction models that process a continuous RGB stream $\{\mathbf I_t\}$ to estimate camera poses and 3D geometry online.
In addition to their constant-memory state design, we emphasize a manifold interpretation: the persistent state can be treated as a compact \emph{subspace representation} of the scene, i.e., a point on a Grassmannian manifold.
Thus, streaming inference induces a latent \emph{trajectory on the Grassmannian}, where a desirable update should preserve temporal coherence by remaining on (or near) the manifold instead of drifting arbitrarily.

Concretely, the standard forward pass consists of three steps:


\noindent\textbf{Tokenization.} Each frame $\mathbf I_t$ is patchified and encoded into visual tokens $\mathbf F_t$ by a shared-weight ViT encoder $\text{Enc}$:
    \begin{equation}
        \mathbf F_t = \text{Enc}(\mathbf I_t).
    \end{equation}
\noindent\textbf{Recurrent State Update and Context Interaction.} The model maintains a persistent state $\mathbf S_{t-1}$ (initialized by learnable tokens).
At time $t$, the interaction module fuses the incoming observation $\mathbf F_t$ with historical context in $\mathbf S_{t-1}$ to produce the updated state $\mathbf S_t$ and output tokens $\mathbf Y_t$:
    \begin{equation}
        [\mathbf S_t,\mathbf Y_t] = \text{Interaction}(\mathbf S_{t-1},\mathbf F_t).
    \end{equation}
\noindent\textbf{Dense Regression.} Task heads (e.g., DPT or MLP) decode $\mathbf Y_t$ into metric-scale 3D pointmaps $\hat{\mathbf X}_t$, confidence masks $\hat{\mathbf N}_t$, and 6-DoF camera poses $\hat{\mathbf P}_t$:
    \begin{equation}
        \{\hat{\mathbf X}_t, \hat{\mathbf P}_t, \hat{\mathbf N}_t\} = \text{Head}(\mathbf Y_t). \label{eq:output_head}
    \end{equation}

While the recurrent formulation is efficient, its state trajectory is typically unconstrained.
Over long horizons, $\mathbf S_t$ can gradually deviate from the manifold implied by coherent scene evolution, manifesting as geometric drift in both structure and pose.




        


\subsection{Revisit Self-Expressive Property}
\label{sec:sec33}

To motivate our regularization, we revisit a classic viewpoint from NRSfM through the lens of the Grassmannian manifold.
A long-standing observation in NRSfM is that shapes in a temporal sequence are not independent; instead, they evolve within low-dimension subspaces, yielding structure constraints across time.
Early methods~\cite{bregler2000recovering} exploited low-rank structure, and Dai~et al.~\cite{dai2014simple} established a standard optimization formulation.
Building on this line, Zhu et al.~\cite{zhu2014complex} proposed a \emph{union-of-subspaces} model to capture richer sequence structure:
\begin{equation}
    \begin{aligned}
        \arg\min_{\mathbf{X}, \mathbf{C}, \mathbf{E}} & \quad \|\mathbf C\|_* + \alpha\|\mathbf{X}\|_* + \|\mathbf{E}\|_l, \\
        \text{s.t.} & \quad \mathbf{X} = \mathbf{X}\mathbf{C}, \label{eq:affinity_matrix} \\
        & \quad \mathbf{W} = \mathbf{M}\mathbf{X}^{\sharp} + \mathbf{E}.
    \end{aligned}
\end{equation}
Where $\mathbf{C} \in \mathbb{R}^{F \times F}$ is the affinity matrix for an $F$-frame sequence. $\|\cdot\|_*$ denote nuclear norm and $\|\cdot\|_l$ denote l2-norm.
Here, $\mathbf X$ is the target shape sequence, $\mathbf E$ is an error term, $\mathbf M$ denotes the projection matrix, and $\mathbf W$ denotes the observation (image sequence or 2D keypoint track) sequence.
The rearrangement $\mathbf{X}^\sharp$ (from~\cite{dai2014simple}) is designed to better impose low-rank constraints.
Crucially, the union-of-subspaces assumption implies that a sequence can be modeled as multiple locally coherent subspaces; in particular, within a short temporal neighborhood $\mathcal{W}_\tau$, the shapes are assumed to lie in a local shape subspace spanned by the shapes in that neighborhood:
\begin{equation}
    \mathbf X_t \in \mathrm{span}\big(\{\mathbf X_i\}_{i\in\mathcal{W}_{\tau}}\big), \quad \forall t \in \mathcal{W}_{\tau}.
\end{equation}

Kumar et al.~\cite{kumar2018scalable} further pointed out that such locally coherent subspaces can be represented as \emph{points on the Grassmannian manifold} $\mathcal{G}(n,r)$, enabling continuous sequences to admit a locally self-expressing structure.
Deng et al.~\cite{deng2022deep} subsequently brought this idea into deep learning by applying self-expressive in a learned latent space.
For a sequence $\mathcal{X} = [\mathbf X_1, \dots,\mathbf X_F]^\top$ with shapes $\mathbf X_i\in\mathbb{R}^{1\times 3P}$, self-expressive states that each element can be reconstructed as a linear combination of others:
\begin{equation}
    \begin{aligned}
        \mathbf{X}_t &= f_{shape}(\text{Enc}(\mathbf I_t)), \\
        \hat{\mathcal{X}} &= g(\mathcal{X}), \quad \mathcal{X} = \{\mathbf X_1,\cdots,\mathbf{X}_F\}, \\
        \text{s.t.} \quad g(\mathcal X) &= f_{dec}(\mathbf{C}\cdot f_{enc}(\mathcal X)),\\
        \mathbf{C} &= H(f_{enc}(\mathcal{X})). \label{eq:reconst_seq}
    \end{aligned}
\end{equation}
Where $\mathbf{C} \in \mathbb{R}^{F \times F}$ is the affinity matrix as in~\autoref{eq:affinity_matrix}.
$\hat{\mathcal X}$ is the regularized output sequence, $f_{shape}(\cdot)$ maps an image to a per-frame shape, $f_{enc}(\cdot)$ embeds the sequence, and $f_{dec}(\cdot)$ decodes it back.

Importantly, this formulation suggests that a model's latent representations can be interpreted as an evolving \textbf{Grassmannian trajectory}. They encode the scene's geometric subspace at each time step.
This observation motivates our design, we can enforce Grassmannian self-expressive as an inference-time constraint on the streaming state trajectory.

\begin{table*}[htbp]
\centering
\setlength{\abovecaptionskip}{0.1cm}
\small
\caption{\textbf{Video depth estimation}. Benchmarking on synthetic (Sintel), dynamic (Bonn), and outdoor (KITTI) sequences. We evaluate the proposed SSR framework against state-of-the-art methods using both scale-invariant and metric depth protocols. Our training-free method demonstrates the state-of-the-art/highly competitive performance across diverse settings. Compared to another training-free method, TTT3R, our approach also exhibits superior performance.}
\label{tab:basic_video}
\resizebox{\textwidth}{!}{
\begin{tabular}{llcccccccc}
\toprule
            & & & & \multicolumn{2}{c}{Sintel} & \multicolumn{2}{c}{Bonn} & \multicolumn{2}{c}{KITTI}    \\ 
            \cmidrule(lr){5-6} \cmidrule(lr){7-8} \cmidrule(lr){9-10}
Alignment & Method & Venue & Online & Abs Rel$~\downarrow$ &  $\delta<1.25~\uparrow$ & Abs Rel$~\downarrow$ &  $\delta<1.25~\uparrow$& Abs Rel$~\downarrow$ &  $\delta<1.25~\uparrow$ \\ \midrule 

\multirow{12}{*}{\shortstack{Per-sequence\\scale}} 
& DUSt3R   & CVPR'24 & \ding{55} & 0.656 & 45.2 & 0.155 & 83.3 & 0.144 & 81.3 \\
& MASt3R   & ECCV'24 & \ding{55} & 0.641 & 43.9 & 0.252 & 70.1 & 0.183 & 74.5 \\
& MonSTR3R & ICLR'25 & \ding{55} & 0.378 & 55.8 & 0.067 & 96.3 & 0.168 & 74.4 \\ 
& Easi3R   & ICCV'25 & \ding{55} & 0.377 & 55.9 & 0.059 & 97.0 & 0.102 & 91.2 \\
& AETHER   & ICCV'25 & \ding{55} & 0.324 & 50.2 & 0.273 & 59.4 & 0.056 & 97.8 \\
& VGGT     & CVPR'25 & \ding{55} & 0.287 & 66.1 & 0.055 & 97.1 & 0.070 & 96.5 \\ \cmidrule{2-10}
& Spann3R  & 3DV'25 & \ding{51} & 0.622 & 42.6 & 0.144 & 81.3 & 0.198 & 73.4 \\
& Point3R  & NeurIPS'25 & \ding{51} & 0.452 & 48.9 & 0.060 & 96.0 & 0.136 & 84.2 \\
& CUT3R    & CVPR'25 & \ding{51} & 0.421 & 47.9 & 0.078 & 93.7 & 0.118 & 88.1 \\
& StreamVGGT& ArXiv'25 & \ding{51} & \textbf{0.323} & \textbf{65.7} & \textbf{0.059} & \textbf{97.2} & 0.173 & 72.1 \\
& TTT3R    & ICLR'26 & \ding{51} & 0.404 & 50.1 & 0.068 & 95.4 & \underline{0.113} & \underline{90.4} \\
& Ours     & Ours & \ding{51} & \underline{0.402} & \underline{50.6} & \underline{0.061} & \underline{96.8} & \textbf{0.109} & \textbf{91.1} \\ \midrule 

\multirow{5}{*}{Metric scale}
& MASt3R   & ECCV'24 & \ding{55} & 1.022 & 14.3 & 0.272 & 70.6 & 0.467 & 15.2 \\
& CUT3R    & CVPR'25 & \ding{51} & 1.029 & 23.8 & 0.103 & 88.5 & 0.122 & 85.5 \\
& Point3R  & NeurIPS'25 & \ding{51} & \textbf{0.777} & 17.1 & 0.137 & \underline{94.7} & 0.191 & 73.8 \\
& TTT3R    & ICLR'26 & \ding{51} & 0.977 & \underline{24.5} & \underline{0.090} & 94.2 & \textbf{0.110} & \textbf{89.1} \\
& Ours     & Ours & \ding{51} & \underline{0.920} & \textbf{25.3} & \textbf{0.068} & \textbf{96.5} & \underline{0.113} & \underline{89.0} \\ \bottomrule
\end{tabular}}
\end{table*}

\subsection{Self-expressive Sequence Regularization}
\label{sec:sec34}

We propose Self-expressive Sequence Regularization (SSR), a training-free correction scheme for streaming 3D reconstruction as~\autoref{fig:pipeline}. 
SSR is motivated by the classic observation that a deforming object evolves through a sequence of states that adhere to \emph{temporal local subspaces}.
When the deformation is smooth over time, consecutive latent states remain close on $\mathcal{G}(n,r)$, a short neighborhood of recent states can be treated as samples from the same local subspace (i.e., a local region of the manifold).
This motivates a \emph{sliding window}: rather than enforcing a global constraint across the entire past, we regularize the trajectory using only the most recent $k$ states, which captures local coherence while keeping memory and computation bounded.

Under this view, SSR imposes \emph{local self-expressiveness}: the current state should be well approximated by a linear combination of its neighbors within the window.
Equivalently, we seek affinity that reconstruct each point from nearby points on $\mathcal{G}(n,r)$.
Concretely, at time $t$ we collect a temporal window $\mathcal{W}_t\in[t-k,\cdots,t]$ with size $k$ and stack states in the temporal window as:
\begin{align}
    \mathcal{S}_t =\mathbf{C}^{(t)}\mathcal{S}_t,\quad \mathcal{S}_t = [\mathbf S_{t-k}, \cdots, \mathbf S_{t}]^\top \label{eq:update_seq}
\end{align}
here, we denote $\mathbf C^{(t)}\in\mathbb{R}^{k\times k}$ as the  affinity matrix with $k$ as the total length of temporal window $\mathcal{S}_t$. In the ideal case (i.e., when the recurrent update remains on a locally consistent trajectory), the state sequence should approximately satisfy the self-expressive relation in~\autoref{eq:update_seq}.

However, in practice the unconstrained update may drift, and $\mathbf S_t$ can deviate from this local subspace structure; SSR therefore computes the  affinity and performs an analytic correction to project the trajectory back toward a locally self-expressive configuration. We compute the affinity matrix $\mathbf{C}$ by measuring similarity. We utilize a mapping function $H(\cdot)$ that encodes feature similarity. For any two frames $i$ and $j$ in temporal group $t$, the  affinity $C_{ij}$ is computed as:

\begin{equation}
    \begin{aligned}
    \mathbf{C}^{(t)} &= H(\mathcal S_t), \\
    C_{ij}^{(t)} &= \frac{\phi(\mathbf{S}_{\mathcal{W}_i}, \mathbf{S}_{\mathcal{W}_j})}{\sum_{k} \phi(\mathbf{S}_{\mathcal{W}_i}, \mathbf{S}_{\mathcal{W}_j})}, \quad i,j \in [0, k]
    \end{aligned}
\end{equation}
where $\phi(\cdot)$ computes similarity by non-normalized dot products. This matrix captures the context of the sequence. As in~\autoref{eq:update_seq}, we reconstruct the state sequence $\mathbf S_t$ using the affinity matrix $\mathbf C^{(t)}$ in a local temporal window $\mathcal{S}_t$.

Given $\mathbf C^{(t)}$, SSR performs an analytic correction by reconstructing states from their neighbors; for example, the corrected state at time $t$ is:
\begin{equation}
    \hat{\mathbf S}_t = \sum_{j=0}^{k}C_{kj}^{(t)} \cdot \mathbf S_{j}.
\end{equation}
Finally, regularizing the state trajectory on $\mathcal{G}(n,r)$ stabilizes the head predictions for both scene geometry and camera motion (\autoref{eq:output_head}). We summarize the full inference-time update in Algorithm~\ref{alg:ssr}.

\begin{algorithm}[htbp]
\caption{Self-expressive Sequence Regularization}
\label{alg:ssr}
\begin{algorithmic} 
\INPUT {$\mathbf I_t,\mathbf S_{t-1}$} \COMMENT{Current image and previous state}
    \STATE $\mathbf F_t \gets \text{Encoder}(\mathbf I_t)$ \COMMENT{Feature Encoding}
    \STATE $[\mathbf S_t, \mathbf Y_t] \gets \text{Interaction}(\mathbf S_{t-1}, \mathbf X_t)$ \COMMENT{Recurrent Update}
    
    \STATE $\mathcal{W}_t \gets [t-k, \dots, t]$ \COMMENT{Maintain temporal window}
    
    \STATE \textbf{compute} $\mathbf{C^{(t)}}$ via similarity:
    \STATE $\mathbf C_{ij}^{(t)} = \frac{\phi(\mathbf S_i, \mathbf S_j)}{\sum_{k} (\phi(\mathbf S_i, \mathbf S_k)}$
    
    \STATE $\hat{\mathcal S}_t \gets C^{(t)} \cdot \mathcal S_t$ \COMMENT{Analytical Correction}
    
\OUTPUT $\hat{\mathbf S}_t, \mathbf Y_t$
\end{algorithmic}

\end{algorithm}

\begin{table*}[t]
\centering
\setlength{\abovecaptionskip}{0.1cm}
\caption{\textbf{Pose estimation}. Overall, our method demonstrates the best comprehensive performance among online methods. On datasets with highly dynamics and long sequences(TUM-D/ScanNet), our method demonstrates superior performance in camera trajectory estimation, reflecting its ability to suppress cumulative drift and adapt to dynamic scenes.}
\label{tab:basic_pose}
\resizebox{\textwidth}{!}{
\begin{tabular}{lccccccccc}
\toprule
            &\multicolumn{3}{c}{Sintel}     & \multicolumn{3}{c}{TUM-dynamics}    & \multicolumn{3}{c}{ScanNet} \\
            \cmidrule(lr){2-4} \cmidrule(lr){5-7} \cmidrule(lr){8-10} 
          Method  & ATE$~\downarrow$     & RPE trans$~\downarrow$ & RPE rot$~\downarrow$ & ATE$~\downarrow$   & RPE trans$~\downarrow$ & RPE rot$~\downarrow$ & ATE$~\downarrow$   & RPE trans$~\downarrow$ & RPE rot$~\downarrow$ \\ \midrule
DUSt3R      & 0.290             & 0.132             & 7.869   & 0.140 & 3.286     & 7.869   & 0.246 & 0.108     & 8.210   \\
MASt3R      & 0.185             & 0.060 & 1.496 & 0.038 & 0.012 & 0.448 & 0.078 & 0.020 & 0.475\\
MonST3R     & 0.111             & 0.044 & 0.869 & 0.098 & 0.019 & 0.935 & 0.077 & 0.018 & 0.529\\
Easi3R      & 0.110             & 0.042 & 0.758 & 0.105 & 0.022 & 1.064 & 0.061 & 0.017 & 0.525\\
AETHER      & 0.189             & 0.054 & 0.694 & 0.092 & 0.012 & 1.106 & 0.176 & 0.028 & 1.204\\
VGGT        & 0.172             & 0.062 & 0.471 & 0.012 & 0.010 & 0.310 & 0.035 & 0.015 & 0.377 \\ \midrule
Spann3R     & 0.329             & 0.110             & 4.471   & 0.056 & 0.021     & 0.591   & 0.096 & 0.023     & 0.661   \\
CUT3R       & 0.213             & \underline{0.066} & \textbf{0.621}   & 0.046 & 0.015     & 0.473   & 0.099 & 0.022     & \underline{0.600}   \\
Point3R     & 0.351             & 0.128             & 1.822   & 0.075  & 0.029 & 0.642 & 0.106 & 0.035 & 1.946 \\
TTT3R       & \textbf{0.201}    & \textbf{0.063}    & 0.671   & \underline{0.028} & \textbf{0.012}     & \underline{0.379}   & 0.064 & \underline{0.021}     & \textbf{0.592}   \\
Ours        & \underline{0.209} & 0.077              & \underline{0.631}   & \textbf{0.026} & \underline{0.013}     & \textbf{0.378}   & \textbf{0.059} & \textbf{0.021}     & 0.721  \\ \bottomrule
\end{tabular}}
\end{table*}

\begin{table*}[htbp]
\centering
\setlength{\abovecaptionskip}{0.1cm}
\caption{\textbf{Sparse view 3D reconstruction}. When input data is extremely sparse and sequence lengths are very short, our method experiences performance degradation to the point where it underperforms baseline methods. This occurs because the proposed method aims to aggregate meaningful information from historical context, whereas such extreme input settings directly contradict this objective. We explore this issue in depth in the analysis section.}
\label{tab:basic_3d}
\begin{tabular}{lllllllllllll}

\toprule
            & \multicolumn{6}{c}{7   Scenes}                                                              & \multicolumn{6}{c}{NRGBD}                                                   \\ \cmidrule(lr){2-7} \cmidrule(lr){8-13} 
            & \multicolumn{2}{c}{$Acc\downarrow$} & \multicolumn{2}{c}{$Comp\downarrow$}        & \multicolumn{2}{c}{$NC\uparrow$}          & \multicolumn{2}{c}{$Acc\downarrow$} & \multicolumn{2}{c}{$Comp\downarrow$} & \multicolumn{2}{c}{$NC\uparrow$} \\ \cmidrule(lr){2-3} \cmidrule(lr){4-5} \cmidrule(lr){6-7} \cmidrule(lr){8-9} \cmidrule(lr){10-11} \cmidrule(lr){12-13}
          Method  & Mean       & Med.       & Mean           & Med.           & Mean           & Med.           & Mean       & Med.       & Mean        & Med.       & Mean       & Med.      \\ \midrule
StreamVGGT  & 0.132      & 0.058      & 0.116          & 0.042          & 0.749          & 0.863          & 0.085      & 0.044      & 0.079       & 0.038      & 0.862      & 0.986     \\
XStreamVGGT & 0.142      & 0.068      & 0.125          & 0.048          & 0.734          & 0.848          & 0.085      & 0.049      & 0.075       & 0.038      & 0.850      & 0.986     \\
DUSt3R-GA   & 0.146      & 0.077      & 0.181          & 0.067          & 0.736          & 0.839          & 0.144      & 0.019      & 0.154       & 0.018      & 0.870      & 0.982     \\
MASt3R-GA   & 0.185      & 0.081      & 0.180          & 0.069          & 0.701          & 0.792          & 0.085      & 0.033      & 0.063       & 0.028      & 0.794      & 0.928     \\
MonST3R-GA  & 0.248      & 0.185      & 0.266          & 0.167          & 0.672          & 0.759          & 0.272      & 0.114      & 0.287       & 0.110      & 0.758      & 0.843     \\
Spann3R     & 0.298      & 0.226      & 0.205          & 0.112          & 0.650          & 0.730          & 0.416      & 0.323      & 0.417       & 0.285      & 0.684      & 0.789     \\
CUT3R       & 0.126      & 0.047      & 0.154 & 0.031 & 0.727 & 0.834 & 0.099      & 0.031      & 0.076       & 0.026      & 0.837      & 0.971     \\
TTT3R       & 0.156      & 0.054      & 0.196          & 0.046          & 0.719          & 0.823          & 0.105      & 0.032      & 0.080       & 0.027      & 0.835      & 0.970     \\
Ours        & 0.132      & 0.047      & 0.167          & 0.031          & 0.724          & 0.832         & 0.100      & 0.036      & 0.078       & 0.032      & 0.835      & 0.970     \\ \bottomrule
\end{tabular}
\end{table*}

\section{Experiments}

In this section, we evaluate the effectiveness of our training-free Self-expressive Sequence Regularization framework across camera pose estimation, video depth estimation, and 3D reconstruction tasks.

\noindent\textbf{Baseline.} We selected CUT3R as our baseline and regularized its state sequences using SSR. To validate the effectiveness of our proposed method, we first conducted foundational quantitative experiments in line with CUT3R. 



\noindent\textbf{Implementation Details.} 
By utilizing a window size (k=8) of historical state features to compute the  affinity matrix $\mathbf{C}$ as Algorithm~\ref{alg:ssr}, the model can outperform the base CUT3R model without the computational overhead of gradient-based updates or test-time training.

\begin{table*}[htbp]
\centering
\setlength{\abovecaptionskip}{0.1cm}
\caption{\textbf{Sequence view 3D reconstruction}. Compared to the content in \autoref{tab:basic_3d}, when we input the same dataset into the proposed method in the form of longer, more continuous sequences, the proposed method achieves better results than the baseline method. This indicates that the performance degradation shown in \autoref{tab:basic_3d}, is primarily caused by the input configuration.}
\label{tab:sequence_3d}
\begin{tabular}{lllllllllllll}
\toprule
            & \multicolumn{6}{c}{7   Scenes}                                                              & \multicolumn{6}{c}{NRGBD}                                                   \\ \cmidrule(lr){2-7} \cmidrule(lr){8-13} 
            & \multicolumn{2}{c}{$Acc~\downarrow$} & \multicolumn{2}{c}{$Comp~\downarrow$}        & \multicolumn{2}{c}{$NC~\uparrow$}          & \multicolumn{2}{c}{$Acc~\downarrow$} & \multicolumn{2}{c}{$Comp~\downarrow$} & \multicolumn{2}{c}{$NC~\uparrow$} \\ \cmidrule(lr){2-3} \cmidrule(lr){4-5} \cmidrule(lr){6-7} \cmidrule(lr){8-9} \cmidrule(lr){10-11} \cmidrule(lr){12-13}
         Method   & Mean       & Med.       & Mean           & Med.           & Mean           & Med.           & Mean       & Med.       & Mean        & Med.       & Mean       & Med.      \\ \midrule
CUT3R       & 0.032      & 0.017      & 0.032          & 0.009          & \textbf{0.641}          & 0\textbf{.720}          & 0.090      & 0.045      & 0.035       & 0.011      & 0.743      & 0.906     \\
Ours        & \textbf{0.024}      & \textbf{0.011}      & \textbf{0.024}          & \textbf{0.006}          & 0.625          & 0.695          & \textbf{0.075}      & \textbf{0.029}      & \textbf{0.029}       & \textbf{0.008}      & \textbf{0.758}      & \textbf{0.914}     \\ \bottomrule
\end{tabular}
\end{table*}

\begin{table}[thb]
\centering
\setlength{\abovecaptionskip}{0.1cm}
\caption{The impact of window length on experimental results exhibits diminishing returns. For typical experiments, we actually do not require excessively long window lengths to obtain satisfactory outcomes. Therefore, the computational overhead associated with SSR can be considered fixed and relatively small.}
\label{tab:ablation_video}
\resizebox{1.0\linewidth}{!}{
\begin{tabular}{lcccc}
\toprule
            & \multicolumn{2}{c}{BONN} & \multicolumn{2}{c}{KITTI}    \\ 
            \cmidrule(lr){2-3} \cmidrule(lr){4-5} 
        Method    & Abs Rel$~\downarrow$ &  $\delta<1.25~\uparrow$ & Abs Rel$~\downarrow$ &  $\delta<1.25~\uparrow$ \\ \midrule
CUT3R       & 0.078 & 93.7 & 0.118 & 88.1 \\
TTT3R	    & 0.064 & 96.2 & 0.110 & 90.6 \\
Ours (k=2)  & 0.064 & 96.6 & 0.112 & 90.2 \\
Ours (k=4)  & 0.062 & 96.8 & 0.109 & \textbf{91.2} \\
Ours (k=16) & 0.061 & 96.8 & 0.111 & 90.6 \\
Ours (k=32) & 0.061 & 96.8 & 0.111 & 90.6 \\
Ours (k=64) & 0.062 & 96.7 & 0.112 & 90.6 \\
Ours        & \textbf{0.061} & \textbf{96.8} & \textbf{0.109} & 91.1 \\ \bottomrule
\end{tabular}}
\end{table}

\subsection{Video Depth Estimation}
Our experiments are based on the setup of CUT3R. We conducted video depth estimation on the KITTI~\cite{Geiger2013IJRR}, Sintel~\cite{Butler:ECCV:2012}, and Bonn~\cite{palazzolo2019iros} datasets. These experiments primarily evaluated the model's ability to perform frame-by-frame depth estimation while maintaining consistency between frames. 
Consistent with CUT3R, our reported scale-invariant relative depth and metric scale absolute depth accuracy. 
We adopt evaluation metrics with both online and offline baselines, specifically Absolute Relative Error (Abs Rel) and $\delta < 1.25$~(the percentage of pixels for which the ratio between the predicted depth $\hat{d}$ and the ground-truth depth $d^*$ is within a factor of 1.25) following~\cite{zhang2024monst3r}.

As shown in~\autoref{tab:basic_video}, the proposed method achieved better results than the baseline model on all three benchmark datasets and also outperformed another untrained method, TTT3R~\cite{chen2025ttt3r}. 
Furthermore, the proposed method did not significantly reduce inference speed despite a slight increase in computational overhead. Notably, the proposed method achieves better performance gains on the longer-sequence Bonn dataset. This demonstrates its capability to preserve and utilize long-range information, enhancing the foundation model's ability to leverage long-term contextual information without requiring training.


\subsection{Pose Estimation}
To determine whether our method can suppress long-range drift issues, we also followed the experimental setup of CUT3R and conducted pose estimation experiments on the TUM-Dynamics~\cite{sturm2012benchmark}, Sintel~\cite{palazzolo2019iros}, and ScanNet~\cite{dai2017scannet} datasets to assess the model's ability to estimate continuous camera trajectories. This task validates the model's ability to infer spatial positions from continuous inputs. We report Absolute Translation Error~(ATE), Relative Translation Error~(RPE Trans), and Relative Rotation Error~(RPE Rot).

As shown in~\autoref{tab:basic_pose}, the proposed method demonstrates significant performance advantages over others across most metrics. Compared to another training-free TTT3R, it also achieves notable improvements. By leveraging self-expressive to construct effective reconstruction constraints, our approach inherently filters dynamic noise on TUM-D, leading to more accurate and reliable depth estimation. For ScanNet, the proposed method also demonstrates improvements, indicating that our approach effectively mitigates information drift caused by long-range information.

\begin{figure*}[tbh]
    \centering
    \includegraphics[width=0.90\textwidth]{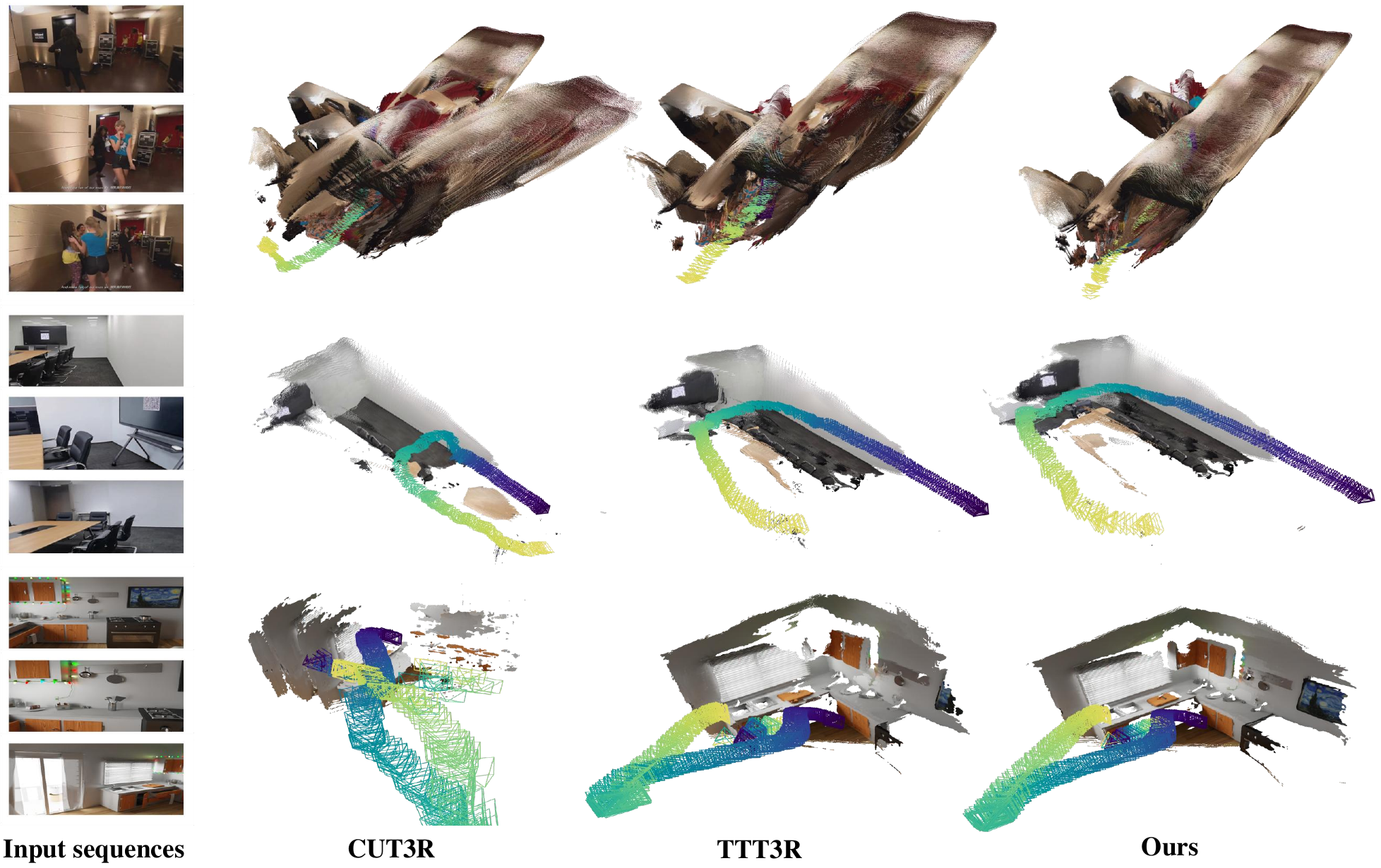} 
    \caption{\textbf{Qualitative Comparison.} As shown in the figure, for long trajectories prone to cumulative drift and for loop trajectories, our method achieves loop closure more effectively than CUT3R and TTT3R, thereby yielding relatively superior reconstruction results.}
    \label{fig:viz_3d}
    \vspace{-4mm}
\end{figure*}

\subsection{3D Reconstrcution}

Since sparse inputs of short length \textbf{contradicts} the core objective of the proposed method, its performance under this configuration warrants our primary analytical focus. To validate the model's scene reconstruction capability from discrete viewpoints, we followed CUT3R's configuration and conducted tests using the 7-Scenes~\cite{shotton2013scene} and NRGBD~\cite{azinovic2022neural} datasets. 
For the 7-Scenes dataset, we selected 3–5 frames per sequence. For the NRGBD dataset, we selected 2–4 frames per sequence. 
In line with~\cite{zhang2024monst3r}, we reported the Accuracy~(ACC), Completion~(Comp), and Normal consistency~(NC) scores. 

As shown in~\autoref{tab:basic_3d}, when confronted with sparse inputs of short length, the proposed method exhibits some performance degradation compared to the baseline. This is primarily because the core idea of the proposed method lies in determining the affinity matrix by computing the self-correlation of historical state sequences. Ideally, when dealing with sparse inputs, the affinity matrix should degenerate into the identity matrix. However, in practice, if inputs are not explicitly timestamped, states separated by large intervals in the historical sequence may still exhibit some degree of similarity. Resulting in an affinity matrix that is close to but not exactly the identity matrix. Consequently, some unrelated states are erroneously merged into the current state, leading to performance degradation.

It is worth noting that TTT3R, which shares a similar conceptual approach to our method (incorporating historical information in some form), exhibited more severe degradation. We speculate this stems from the choice of a different historical information fusion strategy. To further validate whether this degradation stems from inherent dataset issues or, as hypothesized, from differences in input structures, we conducted new experiments in the analysis section.

\subsection{Analysis}

\noindent\textbf{Degradation in Reconstruction.} We first confirmed that this performance degradation was not caused by the data being incompatible with the proposed method. We tested 7-Scenes and the NRGBD dataset with denser continuous sequence (sampling rate of 1/20 on original sequence). 
As shown in ~\autoref{tab:sequence_3d}, 
under this setting, our method achieved better results than the baseline. This indicates that the degradation in performance is not attributable to the distribution of the dataset, but stems from input formats that conflict with the objectives of the proposed method. 

Furthermore, we validated 3D reconstruction capabilities on casual videos. As shown in the visualization results in~\autoref{fig:viz_3d}, the proposed method achieved superior performance for inputs with looping observation and longer trajectories.

\begin{figure}[thb]
    \centering
    \includegraphics[width=0.5\textwidth]{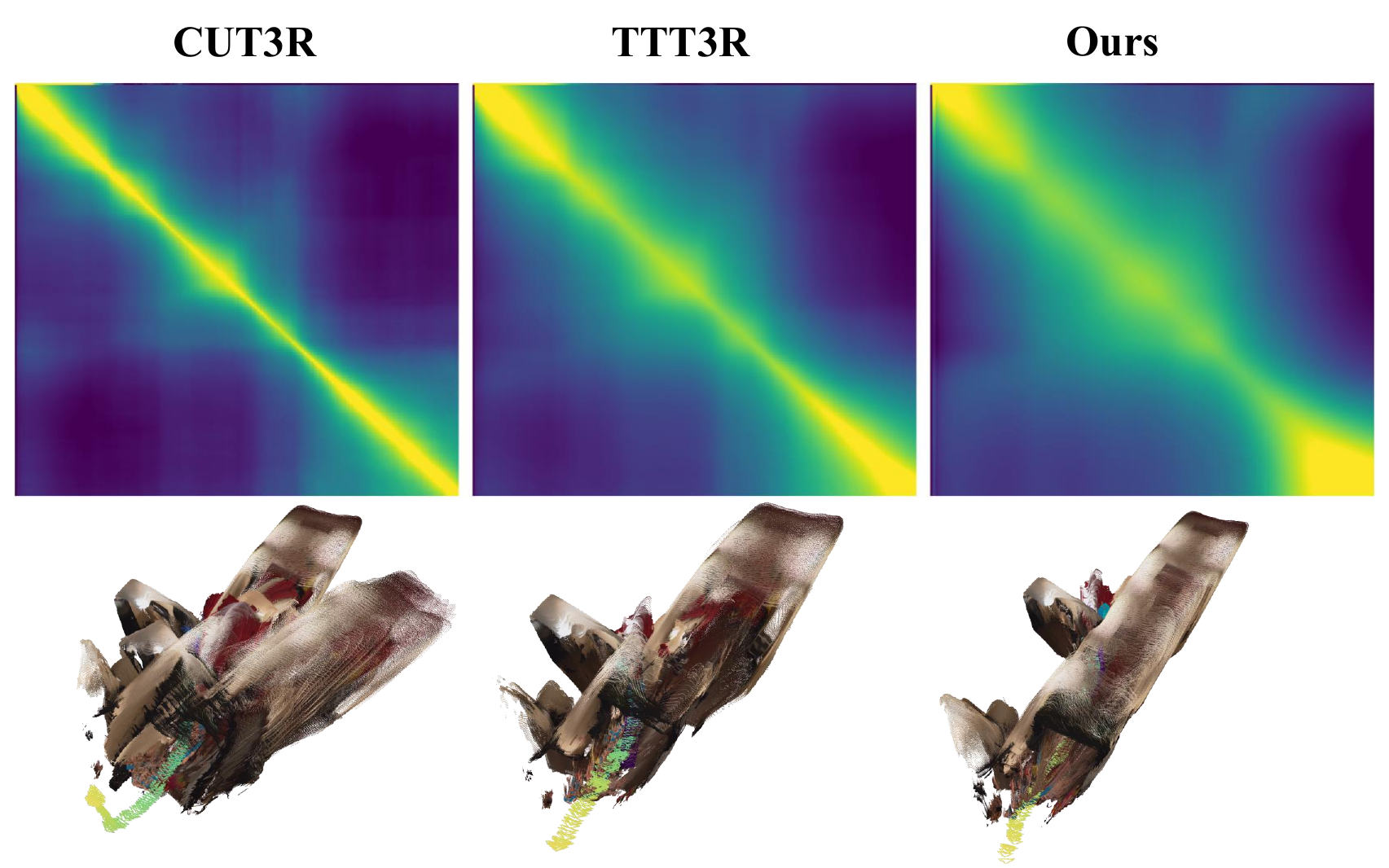} 
    \caption{\textbf{Affinity matrix visualization}. The off-diagonal activations in the affinity matrix clearly indicate that our method maintains strong contextual relationship on early frames even as the sequence progresses, providing a robust context constraint for the current estimation.}
    \label{fig:affinity_comparison}
    \vspace{-5mm}
\end{figure}

\noindent\textbf{Window Size.} Longer window size isn't necessarily better. As shown in~\autoref{tab:ablation_video}, when the sequence exceeds a certain length~(temporal window size 32 as \textit{Ours~(k=32)}), it may not improve results and could even have a negative effect~(temporal window size 64 as \textit{Ours~(k=64)}). As shown in~\autoref{fig:affinity_comparison}, contextual relevance exhibits distinct segmentation patterns, indicating that the size of local subspaces is bounded. Consequently, as demonstrated in the table, a larger window size does not necessarily yield better results.

\noindent\textbf{Affinity Matrix.} Compared to the baseline, both TTT3R and our method demonstrate a superior capacity for leveraging long-range contextual information. As illustrated in~\autoref{fig:affinity_comparison}, our method effectively captures long-distance temporal similarities, which is visually manifested as distinct structures {block} on the affinity heatmap. This capability allows the algorithm to achieve historical states {recall}, thus mitigating cumulative drift. These off-diagonal responses remain stable over time, indicating that the model can retrieve relevant earlier observations rather than relying on adjacent frames.

\noindent\textbf{Limitation.} The most significant limitation of the proposed method has been demonstrated in the preceding discussion. Our approach exhibits detrimental effects on the model when confronted with extremely sparse input data. While this issue can be mitigated through engineering workarounds, it underscores a fundamental shortcoming in our method: the inability to effectively fuse long-range contextual information and to accurately discern the relevance of contextual data to the current context. This represents a key area for future improvement.
\section{Conclusion}



We have proposed a novel Grassmannian manifold perspective to mitigate state drift in streaming 3D reconstruction. 
Our Self-expressive Sequence Regularization (SSR) is a training-free, plug-and-play operator that ensures temporal coherence by constraining state trajectories. 
By exploiting the self-expressive of sequence, SSR suppresses cumulative errors, consistently improving reconstruction precision as demonstrated in our evaluations.

\section{Impact Statement}
This paper presents work whose goal is to advance the field of 3D Vision. There are many potential societal consequences of our work, none which we feel must be specifically highlighted here.

\section*{Acknowledgments}
This research was supported in part by the National Natural Science Foundation of China (62525115, U2570208).

\bibliography{main}
\bibliographystyle{icml2026}

\newpage
\appendix
\onecolumn
\section{Context Forgetting}

Cut3R is a strong foundation model with robust spatial reasoning capabilities and an efficient streaming design for incremental 3D reconstruction. However, its performance is inherently constrained by context forgetting over long horizons. We empirically characterize this limitation through a context-integration experiment, where an attention-like decay operation fuses historical states from previous time steps with the current state. The formulation is:
\begin{equation}
    \hat{\mathbf{S}}_t = \alpha\mathbf{S}_t + (1-\alpha)\mathbf{S}_{t-1}.\quad t\in[0,1]
\end{equation}
Here, the weighting coefficient $\alpha$ governs the trade-off between current-state evidence and historical-state memory when forming the updated representation. Intuitively, smaller values place more emphasis on temporal carry-over, whereas larger values prioritize immediate observations. To quantify how this balance affects reconstruction quality, we evaluate a broad set of $\alpha$ values and report the resulting trends below.

\begin{table}[h]
\centering
\begin{tabular}{cl|cc|cc|cc}

\toprule
\multicolumn{1}{l}{}                                     &                     & \multicolumn{2}{c|}{Sintel}                                            & \multicolumn{2}{c|}{BONN}                                              & \multicolumn{2}{c}{KITTI}                                             \\ 
\cmidrule(lr){3-4} \cmidrule(lr){5-6} \cmidrule(lr){7-8}
\multicolumn{1}{l}{}                                     & Method              & \multicolumn{1}{l}{$Abs Rel~\downarrow$} & \multicolumn{1}{l|}{$\delta~\textless{}1.25~\uparrow$} & \multicolumn{1}{l}{$Abs Rel~\downarrow$} & \multicolumn{1}{l|}{$\delta~\textless{}1.25~\uparrow$} & \multicolumn{1}{l}{$Abs Rel~\downarrow$} & \multicolumn{1}{l}{$\delta~\textless{}1.25~\uparrow$} \\ \midrule
\multicolumn{1}{c|}{\multirow{6}{*}{Per-sequence scale}} & Fuse a=0.1  & 0.426                        & 47.9                                    & 0.076                        & 93.8                                    & 0.122                        & 87.6                                   \\
\multicolumn{1}{c|}{}                                    & Fuse a=0.5  & 0.413                        & 49                                      & 0.733                        & 94.4                                    & 0.118                        & 88.5                                   \\
\multicolumn{1}{c|}{}                                    & Fuse a=0.9  & 0.432                        & 0.496                                   & 0.733                        & 94.4                                    & 0.118                        & 88.5                                   \\
\multicolumn{1}{c|}{}                                    & Cut3R               & 0.421                        & 47.900                                  & 0.078                        & 93.700                                  & 0.118                        & 88.100                                 \\
\multicolumn{1}{c|}{}                                    & TTT3R               & 0.403                        & 49.100                                  & 0.840                        & 94.200                                  & 0.110                        & 90.600                                 \\
\multicolumn{1}{c|}{}                                    & Ours                & 0.401                        & 50.600                                  & \textbf{0.061}               & \textbf{96.800}                         & \textbf{0.109}               & \textbf{91.070}                        \\ \midrule
\multicolumn{1}{c|}{\multirow{6}{*}{Metric scale}}       & Fuse a=0.1   & 1.01                         & 23.6                                    & 0.103                        & 88.6                                    & 0.126                        & 84.2                                   \\
\multicolumn{1}{c|}{}                                    & Fuse a=0.5   & 0.962                        & 23.9                                    & 0.097                        & 90.1                                    & 0.119                        & 86.4                                   \\
\multicolumn{1}{c|}{}                                    & Fuse a=0.9   & 0.917                        & 22.8                                    & 0.097                        & 90.2                                    & 0.119                        & 86.4                                   \\
\multicolumn{1}{c|}{}                                    & Cut3R               & 1.029                        & 23.8                                    & 0.103                        & 88.5                                    & 0.122                        & 85.5                                   \\
\multicolumn{1}{c|}{}                                    & TTT3R               & 0.977                        & 24.5                                    & 0.09                         & 94.2                                    & 0.11                         & 89.1                                   \\
\multicolumn{1}{c|}{}                                    & Ours & 0.92                         & 25.3                                    & 0.068                        & 96.5                                    & 0.113                        & 89                                     \\ \bottomrule
\end{tabular}
\end{table}

These results show that even simple interpolation between consecutive states yields non-trivial gains, leading to two key observations. First, although Cut3R offers a strong reconstruction backbone, its long-horizon performance is clearly limited by context forgetting. Second, the consistent improvements suggest that integrating historical trajectory information is an effective and lightweight way to unlock additional model capacity.

\section{Length Generalization}

As discussed in the main paper, Cut3R is a robust foundation model, but its architecture induces memory decay that limits generalization to longer input sequences. Our method mitigates this issue by systematically injecting historical context, thereby improving performance on extended streams. To evaluate this effect, we conduct experiments across a range of input lengths, with results shown below.

\begin{figure}[h]
    \vspace{-3.5cm}
    \centering
    \includegraphics[width=1.0\textwidth]{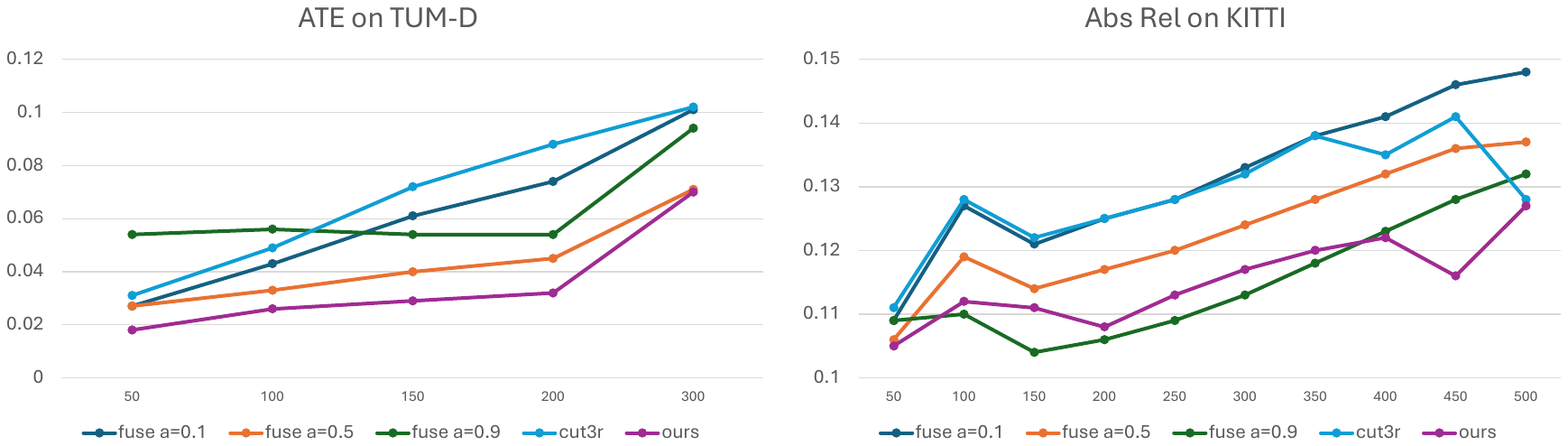} 
    \label{fig:kitti}
        \vspace{-4cm}
\end{figure}

Following the original protocol, the baseline evaluation length for Cut3R on KITTI is set to 110 frames. As sequence length increases, vanilla Cut3R exhibits clear degradation in depth estimation accuracy. In contrast, incorporating the historical context fusion operation consistently improves over the baseline and remains robust across the full range of $\alpha$. The same pattern is observed on TUM-D pose estimation benchmarks. Together, these cross-task and cross-dataset findings support our hypothesis that contextual forgetting is a core limitation of Cut3R, and that reducing this effect is essential for length generalization in streaming spatial models.

\section{Casual Video}

To assess robustness in unconstrained dynamic scenes, we further evaluate on DAVIS, which provides a more in-the-wild setting. The key goal is to examine whether the model can maintain motion continuity, a critical requirement for temporal stability. We therefore benchmark our method on sequences with complex dynamics to test the effectiveness of context integration under challenging conditions.

 \begin{figure}[h]
 \vspace{-3mm}
    \centering
    \includegraphics[width=0.95\textwidth]{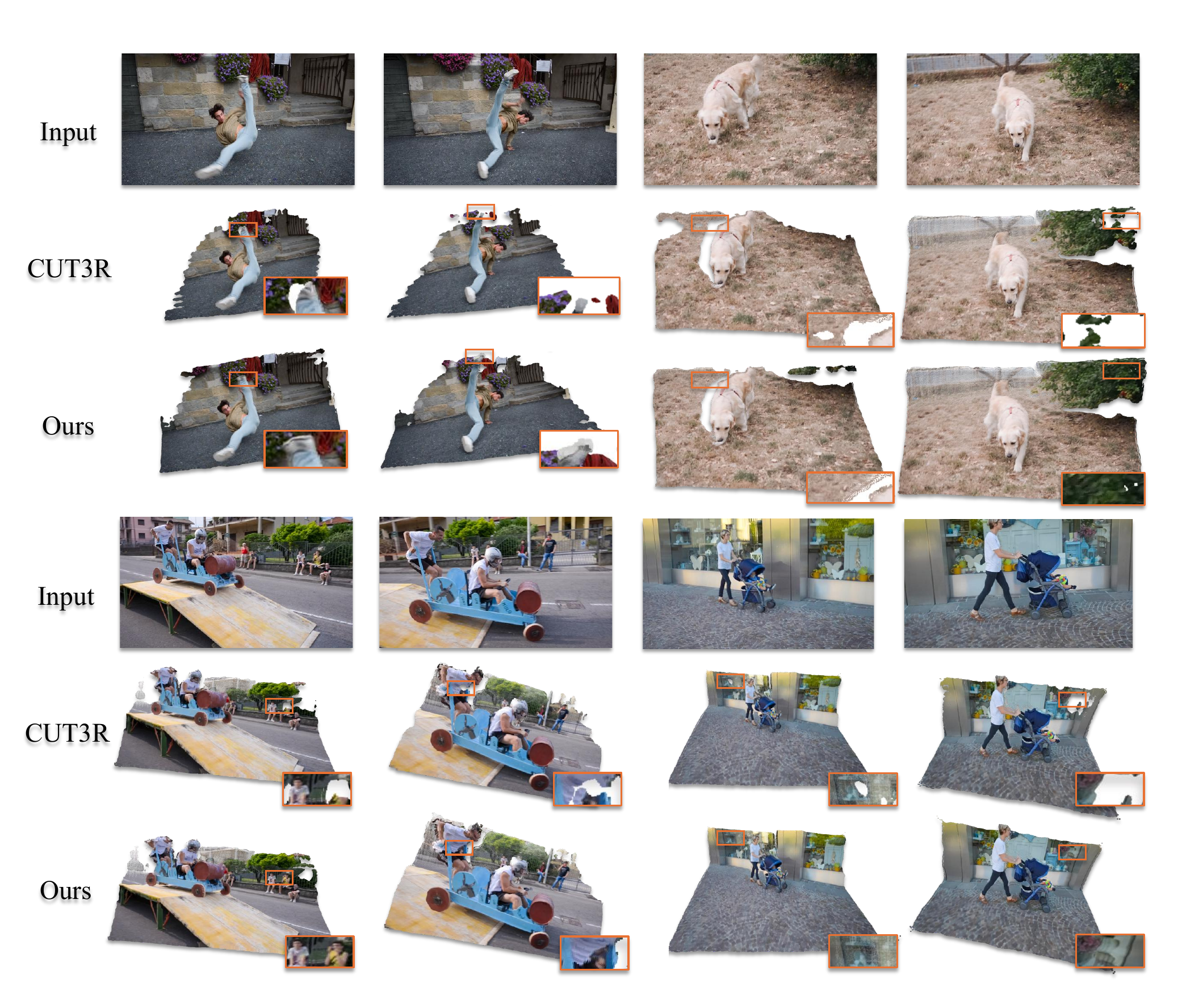} 
    \label{fig:davis}
     \vspace{-1mm}
\end{figure}

Qualitative results show that our approach substantially improves temporal coherence over the foundation model on unconstrained videos, without any additional training. By alleviating long-range contextual decay, the method reduces structural artifacts and prevents catastrophic reconstruction failures in highly dynamic scenes. As a result, it better preserves global scene consistency even under challenging motion patterns.



\end{document}